\documentclass[11pt,a4paper]{article}
\usepackage[hyperref]{acl2018}
\usepackage{times}
\usepackage{latexsym}
\usepackage{graphicx}
\usepackage{times}
\usepackage{url}         
\usepackage{tabularx}
\usepackage{epstopdf}
\usepackage{hyperref}
\usepackage{xstring}

\usepackage{tikzsymbols}
\usepackage{caption,subcaption}

\let\ACMmaketitle=\maketitle
\renewcommand{\maketitle}{\begingroup\let\footnote=\thanks \ACMmaketitle\endgroup}

\usepackage{url}

\usepackage{menukeys}

\usepackage{hologo}

\aclfinalcopy 
\def\aclpaperid{45} 

\title{BCSAT : A Benchmark Corpus for Sentiment Analysis in Telugu Using Word-level Annotations \footnote{* This work was presented at Student Research Workshop in $56^{th}$ Annual Meeting of the Association for Computational Linguistics, ACL. } }

\author{Sreekavitha Parupalli, Vijjini Anvesh Rao and Radhika Mamidi\\
  Language Technologies Research Center (LTRC) \\
  International Institute of Information Technology, Hyderabad \\
  {\tt \{sreekavitha.parupalli, vijjinianvesh.rao\}@research.iiit.ac.in } \\
  {\tt radhika.mamidi@iiit.ac.in} \\}
  
\date{}

\begin{document}
\maketitle
\begin{abstract}
The presented work aims at generating a systematically annotated corpus that can support the enhancement of sentiment analysis tasks in Telugu using word-level sentiment annotations. From OntoSenseNet, we extracted 11,000 adjectives, 253 adverbs, 8483 verbs and sentiment annotation is being done by language experts. We discuss the methodology followed for the polarity annotations and validate the developed resource. This work aims at developing a benchmark corpus, as an extension to SentiWordNet, and baseline accuracy for a model where lexeme annotations are applied for sentiment predictions. The fundamental aim of this paper is to validate and study the possibility of utilizing machine learning algorithms, word-level sentiment annotations in the task of automated sentiment identification. Furthermore, accuracy is improved by annotating the bi-grams extracted from the target corpus. 
\end{abstract}

\section{Introduction}
\label{intro}

Sentiment analysis deals with the task of determining the polarity of text. To distinguish positive and negative opinions in simple texts such as reviews, blogs, and news articles, sentiment analysis (or opinion mining) is used. Over time, it evolved from focusing on explicit opinion expressions to addressing a type of opinion inference which is a result of opinions expressed towards events having positive or negative effects on entities.

There are three ways in which one can perform sentiment analysis : document-level, sentence-level, entity or word-level. These determine the polarity value considering the whole document, sentence-wise polarity, word-wise in some given text respectively  \cite{naidu2017sentiment}. Despite extensive research, the existing solutions and systems have a lot of scope for improvement, to meet the standards of the end users. The main problem arises while cataloging the possibly infinite set of conceptual rules that operate behind the analyzing the hidden polarity of the text \cite{das2011dr}. In this paper, we perform a word-level sentiment annotation to validate the usage of such techniques for improving sentiment analysis task. Furthermore, we use word embeddings of the word-level sentiment annotated lexicon to predict the sentiment label of a document. We experiment with various machine learning algorithms to analyze the affect of word-level sentiment annotations on (document-level) sentiment analysis. 

The paper is organized as follows. In section \ref{sec:relatedwork} we discuss the previous works in the field of sentiment analysis, existing resources for Telugu and specific advances that are made in Telugu.  Section \ref{sec:corpus} describes our corpus and annotation scheme. \ref{sec:experiment} section describes several experiments that are carried out and the accuracies obtained. We also explain the results in detail in \ref{sec:results}. Section \ref{sec:conclusion} showcases our conclusions and section \ref{sec:future} shows the scope for future work. 


\section{Related Work}
\label{sec:relatedwork}

\begin{itemize}

\item{\texttt{Sentiment Analysis:} Several approaches have been proposed to capture the sentiment in the text where each approach addresses the issue at different levels of granularity. Some researchers have proposed methods for document-level sentiment classification \cite{pang2002thumbs,turney2003measuring}. At the top level of granularity, it is often impossible to infer the sentiment expressed about any particular entity, because a document may convey different opinions for different entities. Hence, when we consider the tasks of opinion mining where the sole aim is to capture the sentiment polarities about entities, such as products in product reviews, it has been shown that sentence-level and phrase-level analysis lead to a performance gain \cite{wilson2005recognizing,choi2014+}. In the context of Indian languages, \cite{das2012building} proposes an alternate way to build the resources for multilingual affect analysis where translations into Telugu are done using WordNet. }

\item{\texttt{SentiWordNet :} \cite{das2010sentiwordnet} proposes multiple computational techniques like, WordNet based, dictionary based, corpus based and generative approaches to generate Telugu SentiWordNet. \cite{das2011dr} proposes a tool Dr Sentiment where it automatically creates the PsychoSentiWordNet which is an extension of SentiWordNet that presently holds human psychological knowledge on a few aspects along with sentiment knowledge.
}
 
\item{\texttt{Advances in Telugu:} 
\cite{naidu2017sentiment} utilizes Telugu SentiWordNet on the news corpus to perform the task of Sentiment Analysis. \cite{mukku2017actsa} developed a polarity annotated corpus where positive, negative, neutral polarities are assigned to 5410 sentences in the corpus collected from several sources. They developed a gold standard annotated corpus of Telugu sentences aimed at improving sentiment analysis in Telugu. 
To minimize the dependence of machine learning(ML) approaches for sentiment analysis on abundance of corpus, this paper proposes a novel method to learn representations of resource-poor languages by training them jointly with resource-rich languages using a siamese network \cite{choudhary2018emotions}.
A novel approach to classify sentences into their corresponding sentiment using contrastive learning is proposed by \cite{choudhary2018sentiment} which utilizes the shared parameters of siamese networks.}

\cite{GANGULA18.146} and \cite{mukku2017actsa} are the only reported works for Telugu sentiment analysis using sentence-level annotations who developed annotated corpora. Ours is the first of it's kind NLP research which uses sentiment annotation of bi-grams for sentiment analysis (opinion mining). 

\end{itemize}

\section{Building the Benchmark Corpus}
\label{sec:corpus}

Lexicons play an important role in sentiment analysis. Having annotated lexicon is key to carry out sentiment analysis efficiently. The primary task in sentiment analysis is to identify the polarity of text in any given document. The polarity may be either positive, negative or neutral \cite{naidu2017sentiment}. Sentiment is a property of human intelligence and is not entirely based on the features of a language. Thus, people’s involvement is required to capture the sentiment \cite{das2011dr}. Having said this, we establish that annotated lexicons are of immense importance in any language for  sentiment analysis (a.k.a opinion mining). 

For our experiments, we utilize the reviews dataset from Sentiraama \footnote{\url{https://ltrc.iiit.ac.in/showfile.php?filename=downloads/sentiraama/}} corpus. It contains 668 reviews in total for 267 movies, 201 products and 200 books. Product reviews has 101 positive and 100 negative entries; movie reviews has 136 positive and 132 negative reviews; book reviews data has 100 positive and 100 negative entries. Since the obtained corpus is only annotated with document-level sentiment labels, we perform the word-level sentiment annotation manually. 

\subsection{Annotation Procedure}
In this paper, sentiment polarities are classified into 4 labels : positive, negative, neutral and ambiguous. Positive and negative labels are given in case of positive and negative sentiments in the word respectively. Ambiguous label is given to words which acquire sentiment based on the words it is used along with or it's position in a sentence. Neutral label is given when the word has no sentiment in it. However, neutral and ambiguous sentiment labels are of no significant use for the task of sentiment analysis. Henceforth, those labels are ignored in our experiments. 

Sentiment annotations are performed on two different kinds of data. Table \ref{tab:labels} showcases the distribution of sentiment labels at the word-level.

\begin{itemize}

\item {\texttt{Unigrams:} We obtain 7,663 words from Telugu SentiWordNet \footnote{\url{http://amitavadas.com/sentiwordnet.php}} resource to calculate the base-line accuracy of any word-level sentiment annotated model. These words are already annotated for sentiment/polarity. However, it doesn't provide extensive coverage of Telugu. Later on, we discover a newly developed large resource of Telugu words by \cite{parupalli2018enrichment}, OntoSenseNet, which has a collection of 21,000 words (adjectives+verbs+adverbs). We perform the task of word-level sentiment annotation on the words obtained from this resource and we refer to these annotated words as unigrams throughout this paper. Language experts who performed the annotations are given some guidelines to follow. Experts are implored to look at the word, it's gloss and then decide which one of the four sentiment labels is more apt for a given word. Aforementioned word-level sentiment annotation is an attempt to improve the coverage of SentiWordNet. }

\item {\texttt{Bigrams:} \label{bigraminfo} Furthermore, sentiment cannot always be captured in a single word.This paper aims to check if bigram annotation is a suitable approach for improving the efficiency of sentiment analysis. To validate the hypothesis, we extract bigrams, which occurred at least more than once, only from the target corpus - Sentiraama dataset developed by \cite{GANGULA18.146}. For example, consider the bigram 
('DhokA', 'ledu'). The words individually mean `hurdle (DhokA)', `no (ledu)'. Thus, in word-level annotation task they would be given a negative label. However, the bigram means there is `nothing that can stop' which invokes a positive sentiment. Such occurrences are quite common in the text, especially reviews, which lead us to believe that bigram polarity has potential to enhance sentiment analysis, opinion mining. The usage of this developed resource in experiments performed is explained in section \ref{sec:experiment}. }

\end{itemize}
  
\subsection{Validation}
Annotations are done by 2 native speakers of Telugu. If the annotators aren't able to decide which label to assign, they are advised to tag it as uncertain. In case of a disagreement, the label given by the annotator with more experience is given priority. Validation of the developed resource is done using Cohen's Kappa \cite{cohen1968weighted}. By considering the uncertain cases as borderline cases (where at least one annotator tagged the word as uncertain), Kappa value is seen as 0.91. This shows almost perfect agreement and this proves the consistency in annotation task. This is especially high because when both the annotators are uncertain, we did a re-iteration to finalize the tag. Such re-iterative task is done for about 2,400 words during the development of our resource.  

\begin{table*}
\begin{tabular}{| c | c  c  c  c  c|}
\hline
\textbf{Resource}&\textbf{Positive}&\textbf{Negative}&\textbf{Neutral}&\textbf{Ambiguous}&\textbf{Total}\\
\hline
SentiWordNet \footnote{\url{http://amitavadas.com/sentiwordnet.php}} &2135 & 4076 & 359&  1093 & 7663\\
Dictionary \cite{parupalli2018enrichment} & 3080 & 4232 & 3391 & 10199 & 20896\\
Bigrams & 1978 & 1762 & 8990  & 1996 &14826\\
\hline
\end{tabular}
\caption{Distribution of Sentiment Labels in Several Resources}
\label{tab:labels}
\end{table*}

\section{Experiments and Results}
\label{sec:experiment}
In this section we will analyze and observe how word-level polarity affects overall sentiment of the text through majority polling approach and machine learning based classification approaches. 

\subsection{Majority Polling Approach}
\label{sec:exp1}
A simple intuitive approach to identify the sentiment label of the text is to calculate the sum of positive(+1) and negative(-1) polarity values in it. If the sum is positive, it shows that number of positive words have outnumbered the number of negative words thus resulting in a positive sentiment on the whole. Otherwise, the polarity of the text is negative. Cases where the sum equals to 0 are ignored. 
Following are the word-level polarities we consider for positive and negative labels:
\begin{itemize}
\item \texttt{Unigram}: We use the annotated unigram data that is discussed in \ref{sec:corpus}. For each review, we consider the unigram labels to carry the majority polling approach. 
\item \texttt{Bigram}: The extracted bigrams are annotated for positive and negative polarity. Initially, we divide our data into training and testing sets in 7:3 ratio. We only consider the annotated bigrams from the training corpus to predict the sentiment polarity of reviews in the test data.  
\item \texttt{Unigram+Bigram}: In this trial, we combine the unigram and bigram data to perform majority polling. We consider the whole unigram data whereas bigrams extracted from the training set are only considered for predictions. 
\end{itemize}

Furthermore, as Telugu is agglutinative in nature \cite{pingali2006hindi}, we experiment with the above mentioned approaches after performing morphological segmentation provided by Indic NLP library \footnote{\url{http://anoopkunchukuttan.github.io/indic_nlp_library/}}. Morphological segmentation is performed on the original reviews data and n-grams (positive and negative labels) to see if we could get more accurate sentiment prediction of the reviews due to increment in the coverage. 

\subsection{Machine Learning Based Classification Approach}
In this section, we perform document-level sentiment analysis task with word embedding models, specifically Word2Vec. We utilize a Word2Vec model that is trained on corpus consisting of scrapped data from Telugu websites, with 270 million non-unique tokens on the whole. Furthermore, to obtain vectors for each review, we take word vector of every word in the review and calculate their average to get a single document vector.

\begin{figure}[!h]
\begin{center}
\includegraphics[scale=0.3]{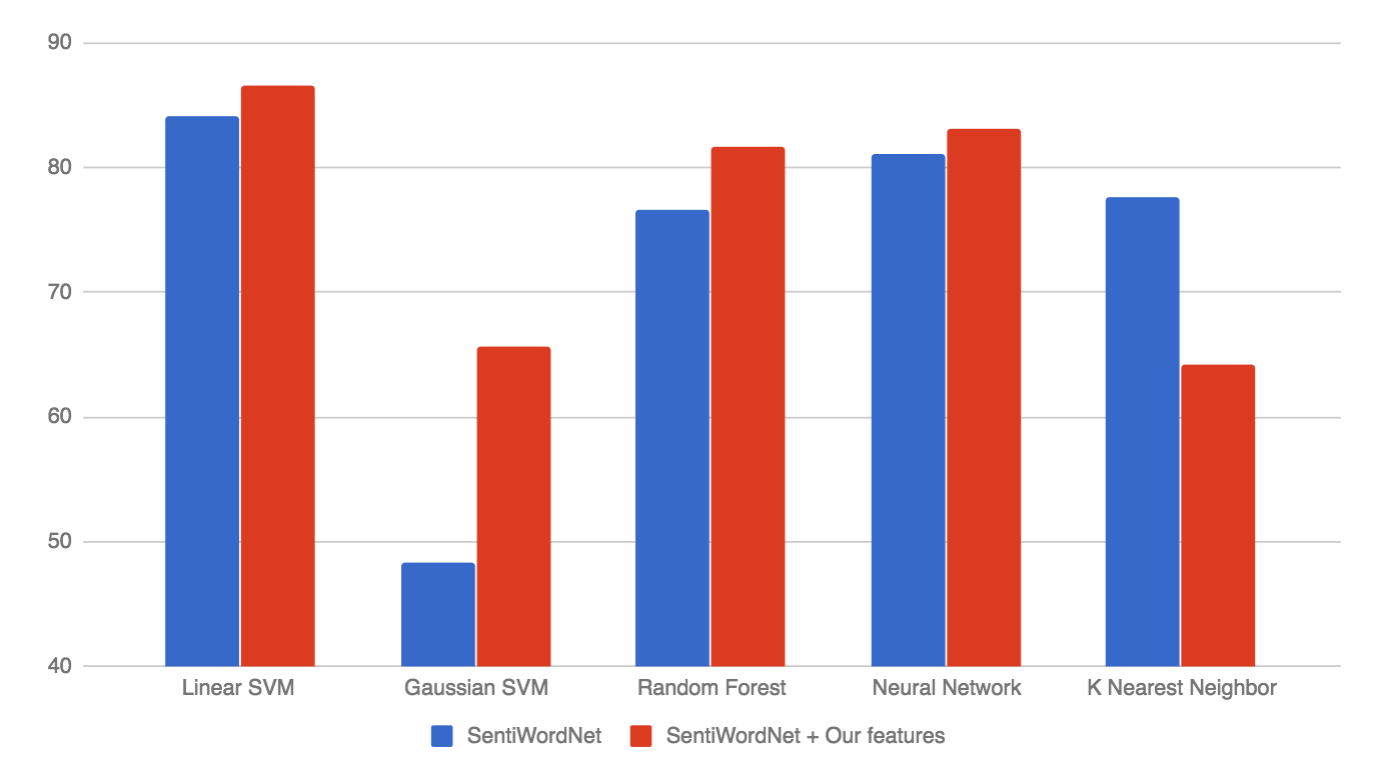} 
\caption{Comparative analysis of percentage accuracies produced by various classifiers}
\label{fig1}
\end{center}
\end{figure}

Though traditional vector-based word representations help us accomplish various natural language processing tasks, they often lack information related to sentiment analysis. Thus, we aim to enrich the Word2Vec vectors obtained from corpus by incorporating word-level polarity features. We do this by adding the features we propose in \ref{sec:exp1} to the original averaged Word2Vec vector, which is expected to increase the accuracy of polarity prediction. The additional features we added are: \textit{positive unigrams} (number of positive polarity unigrams in the review), \textit{negative unigrams} (number of negative polarity unigrams in the review), \textit{positive bigrams }(number of positive polarity bigrams in the review), \textit{negative bigrams} (number of negative polarity bigrams in the review). We partition these document vectors into training and testing sets to develop various classifier models. In this paper, we have implemented 5 classifiers, namely, Linear SVM, Gaussian SVM, Random Forest, Neural Network, K Nearest Neighbor (KNN). Percentage accuracies are illustrated along with the improvement in accuracies after addition of our proposed features in Figure \ref{fig1} and results are discussed in \ref{sec:results}. 


\begin{table*}
\begin{center}
\begin{tabular}{| c | c  c  c  c |}
\hline
\textbf{}&\textbf{SentiWordNet}&\textbf{Our resource}&\textbf{Bigram}&\textbf{Uni+Bigrams}\\
\hline
Before Segmentation & 61.86 & 62.84 &78.97 &55.44 \\
Unclassified reviews & 23/201 & 14/201 & 108/201 & 10/201 \\

\hline
After Segmentation & 60.23 & 58.29 & 49.46 & 57.89 \\
Unclassified reviews & 20/201 & 18/201 & 36/201	& 8/201 \\
\hline
\end{tabular}
\end{center}
\caption{Comparison of accuracies obtained through majority polling on different resources.}
\label{tab:polling}
\end{table*}

\subsection{Results}
\label{sec:results}
In this section, we showcase and analyze the results of the two experiments we have done in section \ref{sec:experiment}. 

\subsubsection{Majority Polling Approach :} Results illustrated in Table \ref{tab:polling} show that certain word-level features do capture information relevant to document-level sentiment analysis. Our hypothesis in Section \ref{bigraminfo} shows that bigram polarity annotations have potential to enhance sentiment analysis. High accuracy obtained by using only bigrams for majority polling proves our hypothesis. However, there is a trade-off between coverage and accuracy. This can be depicted from the huge increase in the count of unclassified reviews in case of bigram majority poling. We also observe that effect of morphological segmentation on accuracy is hardly positive. This indicates that in case of Telugu, morphological data has relevance to sentiment expressed and morphological segmentation would result in loss of such valuable information for sentiment analysis tasks. 
\subsubsection{Machine Learning Based Classification Approach:} This approach shows that across all the classifiers, addition of word-level polarity features improves the process of classification. Therefore, classifiers can predict document-level sentiment polarity with better accuracies. Hence, our hypothesis is validated once again. Accuracies doesn't improve significantly over the baseline value but show a small increment always. KNN classifier shows a huge drop in accuracy after inclusion of the new features proposed. This is observed because KNN assumes all features to hold equal importance for classification. Hence, KNN fails to ignore the noisy features which explains the drop. Random forest and neural network classifiers don't show significant learning from the proposed features. Finally, we observe that linear SVM classifier works best to identify the polarity of a text for our features indicating linear separability of the data. This also explains the bad performance of Gaussian SVM. Linear SVM produces an accuracy of 84.08\% when SentiWordNet words alone are used as a feature, which can be considered as the baseline accuracy. It gives an accuracy of 83.44\% ,84.34\% and 86.57\% for unigrams, bigrams and unigrams+bigrams respectively as features of Linear SVM classifier. 

\section{Conclusions}
\label{sec:conclusion}

In this paper, efforts are made to develop an annotated corpus of 21,000 words to enrich Telugu SentiWordNet. This is a work in progress. We perform annotations of 14,000 bigrams that are extracted from target corpus to validate their importance. This is a first-of-it's-kind approach in Telugu to enhance sentiment analysis. Manual annotations done show perfect agreement which validates the developed resource. Furthermore, we provide a justification to why word-level sentiment annotation of bigrams enhances sentiment analysis though an intuitive majority polling approach, by using several ML classifiers. The results are analyzed for further insights.

\section{Future Work}
\label{sec:future}

We extract bigrams only from the target corpus because we wanted to mainly validate the importance of bigrams in sentiment analysis. However, attempts should be made to enhance the SentiWordNet with, at least, some most occurring bigrams in Telugu. We hope this corpus can serve as a basis for more work to be done in the area of sentiment analysis for Telugu. A continuation to this paper could be handling the enrichment of adjectives and adverbs available in OntoSenseNet for Telugu.

\subsection{Crowd sourcing}
We can develop a crowd sourcing platform where the annotations can be done by several language experts instead of a few. This helps in the annotation of large corpora. We aim to develop a crowd sourcing model for the same in near future. This would be of immense help in annotation of 21,000 unigrams extracted from the dictionary developed by \cite{parupalli2018enrichment}. 

\section{Acknowledgements}
This work is part of the ongoing MS thesis in Exact Humanities under the guidance of Prof. Radhika Mamidi. I am immensely grateful to Vijaya Lakshmi for helping me with data collection. I would like to thank Nurendra Choudary for reviewing the paper and for his part in the ideation. I would like to extend my gratitude to Abhilash Reddy for annotating the dataset, reviewing the work carefully and constantly pushing us to do better. I want to thank Rama Rohit Reddy for his support and for validating the novelty of this research at several points. I acknowledge the support of Google in the form of an International Travel Grant, which enabled me to attend this conference.

\bibliography{acl2018}

\begin{thebibliography}{15}
\expandafter\ifx\csname natexlab\endcsname\relax\def\natexlab#1{#1}\fi

\bibitem[{Choi and Wiebe(2014)}]{choi2014+}
Yoonjung Choi and Janyce Wiebe. 2014.
\newblock +/-effectwordnet: Sense-level lexicon acquisition for opinion
  inference.
\newblock In \emph{Proceedings of the 2014 Conference on Empirical Methods in
  Natural Language Processing (EMNLP)}, pages 1181--1191.

\bibitem[{Choudhary et~al.(2018{\natexlab{a}})Choudhary, Singh, Bindlish, and
  Shrivastava}]{choudhary2018emotions}
Nurendra Choudhary, Rajat Singh, Ishita Bindlish, and Manish Shrivastava.
  2018{\natexlab{a}}.
\newblock Emotions are universal: Learning sentiment based representations of
  resource-poor languages using siamese networks.
\newblock \emph{arXiv preprint arXiv:1804.00805}.

\bibitem[{Choudhary et~al.(2018{\natexlab{b}})Choudhary, Singh, Bindlish, and
  Shrivastava}]{choudhary2018sentiment}
Nurendra Choudhary, Rajat Singh, Ishita Bindlish, and Manish Shrivastava.
  2018{\natexlab{b}}.
\newblock Sentiment analysis of code-mixed languages leveraging resource rich
  languages.
\newblock \emph{arXiv preprint arXiv:1804.00806}.

\bibitem[{Cohen(1968)}]{cohen1968weighted}
Jacob Cohen. 1968.
\newblock Weighted kappa: Nominal scale agreement provision for scaled
  disagreement or partial credit.
\newblock \emph{Psychological bulletin}, 70(4):213.

\bibitem[{Das and Bandyopadhyay(2010)}]{das2010sentiwordnet}
Amitava Das and Sivaji Bandyopadhyay. 2010.
\newblock Sentiwordnet for indian languages.
\newblock In \emph{Proceedings of the Eighth Workshop on Asian Language
  Resouces}, pages 56--63.

\bibitem[{Das and Bandyopadhyay(2011)}]{das2011dr}
Amitava Das and Sivaji Bandyopadhyay. 2011.
\newblock Dr sentiment knows everything!
\newblock In \emph{Proceedings of the 49th annual meeting of the association
  for computational linguistics: human language technologies: systems
  demonstrations}, pages 50--55. Association for Computational Linguistics.

\bibitem[{Das et~al.(2012)Das, Poria, Dasari, and
  Bandyopadhyay}]{das2012building}
Dipankar Das, Soujanya Poria, Chandra~Mohan Dasari, and Sivaji Bandyopadhyay.
  2012.
\newblock Building resources for multilingual affect analysis--a case study on
  hindi, bengali and telugu.
\newblock In \emph{Workshop Programme}, page~54.

\bibitem[{Gangula and Mamidi(2018)}]{GANGULA18.146}
Rama Rohit~Reddy Gangula and Radhika Mamidi. 2018.
\newblock Resource creation towards automated sentiment analysis in telugu (a
  low resource language) and integrating multiple domain sources to enhance
  sentiment prediction.
\newblock In \emph{Proceedings of the Eleventh International Conference on
  Language Resources and Evaluation (LREC 2018)}, Paris, France. European
  Language Resources Association (ELRA).

\bibitem[{Mukku and Mamidi(2017)}]{mukku2017actsa}
Sandeep~Sricharan Mukku and Radhika Mamidi. 2017.
\newblock Actsa: Annotated corpus for telugu sentiment analysis.
\newblock In \emph{Proceedings of the First Workshop on Building Linguistically
  Generalizable NLP Systems}, pages 54--58.

\bibitem[{Naidu et~al.(2017)Naidu, Bharti, Babu, and
  Mohapatra}]{naidu2017sentiment}
Reddy Naidu, Santosh~Kumar Bharti, Korra~Sathya Babu, and Ramesh~Kumar
  Mohapatra. 2017.
\newblock Sentiment analysis using telugu sentiwordnet.

\bibitem[{Pang et~al.(2002)Pang, Lee, and Vaithyanathan}]{pang2002thumbs}
Bo~Pang, Lillian Lee, and Shivakumar Vaithyanathan. 2002.
\newblock Thumbs up?: sentiment classification using machine learning
  techniques.
\newblock In \emph{Proceedings of the ACL-02 conference on Empirical methods in
  natural language processing-Volume 10}, pages 79--86. Association for
  Computational Linguistics.

\bibitem[{Parupalli and Singh(2018)}]{parupalli2018enrichment}
Sreekavitha Parupalli and Navjyoti Singh. 2018.
\newblock Enrichment of ontosensenet: Adding a sense-annotated telugu lexicon.
\newblock \emph{arXiv preprint arXiv:1804.02186}.

\bibitem[{Pingali and Varma(2006)}]{pingali2006hindi}
Prasad Pingali and Vasudeva Varma. 2006.
\newblock Hindi and telugu to english cross language information retrieval at
  clef 2006.
\newblock In \emph{CLEF (Working Notes)}.

\bibitem[{Turney and Littman(2003)}]{turney2003measuring}
Peter~D Turney and Michael~L Littman. 2003.
\newblock Measuring praise and criticism: Inference of semantic orientation
  from association.
\newblock \emph{ACM Transactions on Information Systems (TOIS)},
  21(4):315--346.

\bibitem[{Wilson et~al.(2005)Wilson, Wiebe, and
  Hoffmann}]{wilson2005recognizing}
Theresa Wilson, Janyce Wiebe, and Paul Hoffmann. 2005.
\newblock Recognizing contextual polarity in phrase-level sentiment analysis.
\newblock In \emph{Proceedings of the conference on human language technology
  and empirical methods in natural language processing}, pages 347--354.
  Association for Computational Linguistics.

\end{thebibliography}
\bibliographystyle{acl_natbib}

\end{document}